\documentclass{article}

\usepackage{PRIMEarxiv}

\usepackage[utf8]{inputenc} 
\usepackage[T1]{fontenc}    
\usepackage{hyperref}       
\usepackage{url}            
\usepackage{booktabs}       
\usepackage{amsfonts}       
\usepackage{nicefrac}       
\usepackage{microtype}      
\usepackage{lipsum}
\usepackage{fancyhdr}       
\usepackage{graphicx}       
\usepackage{xcolor}
\graphicspath{{media/}}     
\newtheorem{thm}{\textbf{Theorem}}

\title{On the properties of Gaussian Copula Mixture Models 
\thanks{\textit{\underline{Citation}}: 
\textbf{Ke Wan, Alain Kornhauser, On the properties of Gaussian Copula Mixture Models}} 
}

\author{
  Ke Wan \\
  Dr. \\
  Princeton University \\
  Princeton\\
  \texttt{kwan@alumni.princeton.edu} \\
   \And
  Alain Kornhauser \\
  Professor \\
  Princeton University \\
  Princeton\\
  \texttt{alaink@princeton.edu} \\
}

\begin{document}
\maketitle

\begin{abstract}
  This paper investigates Gaussian copula mixture models (GCMM), which are an extension of Gaussian mixture models (GMM) that incorporate copula concepts. The paper presents the mathematical definition of GCMM and explores the properties of its likelihood function. Additionally, the paper proposes extended Expectation Maximum algorithms to estimate parameters for the mixture of copulas. The marginal distributions corresponding to each component are estimated separately using non parametric statistical methods. In the experiment, GCMM demonstrates improved goodness-of-fitting compared to GMM when using the same number of clusters. Furthermore, GCMM has the ability to leverage un-synchronized data across dimensions for more comprehensive data analysis.

  \end{abstract}
  

  \section{Introduction}
  
  Gaussian Mixture models have been employed in various areas of
  research (Yang 1998 [13] and Pekka 2006 [8]). In the present study,
  we extend Gaussian Mixture Models into Gaussian Copula Mixture
  Models to address the following two concerns:
  \begin{itemize}
   \item  Heavy-tailed data require
  increasing numbers of clusters to fit with GMMs. To control number
  of clusters, heavy tails on marginal distributions should not lead
  to significantly greater clusters given the same underlying
  dependence structure.
  \item GMMs are usually
  applied to a synchronized data matrix of dimension $M$ and number of
  observations $N$. In many problems, there are numerous
  unsynchronized data each dimension, the number of which is denoted
  as $n_m$ for the $m$-th dimension. Such data should be utilized to
  update the joint distribution shared by the different dimensions.
  \end{itemize}
  
  To address the concerns, we introduced copulas into mixture models
  and new Expectation Maximum type algorithms are developed to
  estimate their parameters.
  
  \section{Related Studies}
  Gaussian mixture models have been used widely in various
  applications and the Expectation Maximum algorithm has been utilized
  for estimating their parameters. The convergence properties of such
  Expectation Maximum algorithms have been discussed in Lei (1996
  [12]). However, each component of a GMM is a multivariate gaussian
  distribution that cannot effectively capture heavy tails and the
  number of components become sensitive w.r.t heavy tails. The
  introduction of more flexible components may help to further reduce
  number of components when working with heavy-tailed data.
  
  On the other hand, copulas have been used in research for model
  dependence. The definition of a copula in the two dimensional case
  is given as below:
  
  Let $P$ be a conditional bivariate distribution function with
  continuous margins $F_X$ and $F_Y$,
   and let $\mathcal{F}$ be some conditioning set. There then exists a unique conditional copula
  $C:[0,1]\times[0,1]$ such that (Sklar (1959) [9]):
  \begin{eqnarray}
  P(x,y|F)&=&C(F_X(x|\mathcal{F}),F_Y(y|\mathcal{F})|\mathcal{F}),
  \forall x, y \in R
  \end{eqnarray}
  
  The definitions above can easily be generalized to higher
  dimensions. The advantages of the copula method include the
  following:
  \begin{itemize}
  \item Heavy-tailed joint distributions can be modeled;
  \item Marginal distributions
  and their dependence structure can be studied separately;
  \item Copulas can be calibrated to data sets that are sparse and unevenly distributed.
  \end{itemize}
  
  Upper tail dependence can be studied using copulas (Nelson 2006 [7])
  and copulas can be estimated using a two-step maximum likelihood
  method the properties of which are discussed in White (1994 [10]).
  In the two dimensional case, Archimedean copulas such as BB1 are
  more flexible than Gaussian in capturing heavy tails while the
  estimation of higher dimensional Archimedean copulas may not be as
  fully studied as in the two dimensional case (Marius 2012 [6]).
  Factor models have been introduced to control model complexity as
  well (DongHwan 2011 [4]). Within this context, a mixture of Gaussian
  copulas presents an effective alternative method for improving model
  performance if one wants to study complex dependence structures
  based on simple copulas.
  
  Finally, Gaussian Copula Mixture Models are developed to meet both
  needs. Gaussian Copula Mixture Models can be viewed as extension of
  Gaussian Mixture Models ([8]] and [13]),
  which aim to address the following two concerns:
  \begin{itemize}
   \item  Heavy-tailed data require
  increasing numbers of clusters to fit GMMs. To capture the control
  number of clusters, heavy tails on marginal distributions may lead
  to greater number of clusters in GMM. However, if the heavy-tailed
  data appear independently on each dimension, we should not use
  increasing number of clusters to describe them; in another word,
  multidimensional cluster should be introduced in copula space
  instead of the original data space and heavy tailed marginal
  distributions should be modeled separately. These intuition leads to
  GCMM, in which marginal distributions can be updated using
  non-parametric methods, and mixture models are used to model the
  dependent structure. Such a model potentially leads to fewer number
  of clusters.
  \item GMMs are usually
  applied to a synchronized panel data matrix of dimension $M$ and
  number of observations $N$. In many problems, there are numerous
  unsynchronized data on each dimension, the number of which is
  denoted as $n_m$ for the $m$-th dimension. Such data should be
  utilized to update the joint distribution shared by the different
  dimensions. For a concrete example, if we have 500 observations on
  variable A and 400 observations on variable B, with 300 by 2
  observations  which are synchronized data between A and B, GMM will
  utilize the 300 by 2 observations to update the mixture model while
  GCMM can utilize 300 by 2 observations points to update the mixture
  copula structure. But GCMM will further utilize the unsynchronized
  200 observations for A and 100 observations for B to update their
  marginal distributions respectively, which further contributes to
  the estimation of the copula mixture during iteration.
  \end{itemize}

Ke [14] proposed implicit Gaussian mixture models in 2010 and summarized its theoretical properties in the PHD dissertation as in 2014 [15]. Gaussian copula mixture models are extension to GMM and expectation maximum method was used to generate estimates for the joint distribution of travel time on nearby highways. This paper extends the PHD dissertation and discussed the theoretical properties of the Gaussian copula mixture models and proposed ways to employed usage of un-synchronized data in the EM algorithm. Such theoretical study provided foundations for all relevant applications on different data set.

Independently there is a similar term called Gaussian Mixture Copula Models which was introduced by Tewari 2011 [16], where EM method and gradient descent method was proposed to estimate the distributions. However, the theoretical properties of the log likelihood is not fully explored and how marginal data can be explored in the estimation process can be further studied. Rajan 2016 [17] used Gaussian mixture copulas, to model complex dependencies beyond those captured by meta–Gaussian distributions, for clustering. Bilgrawu 2016 [18] presented and discussed an improved implementation in R of both classes of GMCMs along with various alternative optimization routines to the EM algorithm. Kasa 2020 [19] real high-dimensional gene-expression and clinical data sets showed that HD-GMCM outperforms state-of-the-art model-based clustering methods, by virtue of modeling non-Gaussian data and being robust to outliers through the use of Gaussian mixture copula. Sheikholeslami 2021 [20] uses Gaussian mixture copulas to approximate the joint probability density function of a given set of input-output pairs for estimating the variance-based sensitivity indices.

On Bayesian stats side, Feldman 2022 [21] developed a novel Bayesian mixture copula for joint and non-parametric modeling of multivariate count, continuous, ordinal, and unordered categorical variables. In Zou 2022 [22], a high-dimensional Vine-Gaussian mixture Copula model is combined with Bayesian CNN-BiLSTM model to evaluate uncertainties of model output.

  \section{Mathematical Definitions}
  
   A Gaussian copula mixture model (GCMM) consists
  of a weighted sum of a finite number of joint distributions, each of
  which contains a Gaussian copula. It is a generalization of the
  usual a Gaussian mixture model (GMM). When the marginal
  distributions are restricted to be Gaussian, the model reduces to a
  GMM. To begin, the multivariate Gaussian copula is defined by the
  following probability function:
  
  \begin{eqnarray}
  F(u|P)=\int_{-\infty}^{\Psi^{-1}(u_{1})}...\int_{-\infty}^{\Psi^{-1}(u_{d})}\frac{1}{(2\pi)^{n/2}P^{1/2}}exp(-\frac{1}{2}v^TP^{-1}v)dv
  \end{eqnarray}
  
   whose density is given by
  \begin{eqnarray}
  f(u|P)=\frac{1}{(2\pi)^{n/2}P^{1/2}}exp(-\frac{1}{2}u^TP^{-1}u)\prod_{d=1}^D\frac{1}{\frac{1}{\sqrt{2\pi}}exp(-\frac{1}{2}(\Psi^{-1}(u_d))^2)}
  \end{eqnarray}
  
  where \begin{itemize}
   \item $\Psi$ is the one dimensional cumulative distribution function for a standard normal distribution with density $\psi$;
  \item  $P$ is the copula parameter matrix;
   \item $d$ is the number of dimension.
  \end{itemize}
  Then, with the Gaussianlization of original data on each dimension,
  a GCMM for the joint distribution of a random vector $X$ can be
  defined as follows:
  
  \begin{eqnarray}
  F(X|\pi)=\sum_{k=1}^{K}\pi_k\int_{-\infty}^{Y_{k1}}...\int_{-\infty}^{Y_{kd}}\frac{1}{(2\pi)^{n/2}P_k^{1/2}}exp(-\frac{1}{2}Y^TP_kY)dY
  \end{eqnarray}
   where
  \begin{itemize}
  \item $x=[x_1\ldots x_d]$ is the marginal observation.
  \item $Y_{k}=[Y_{1d} \ldots Y_{kd}]$ is the vector of the transferred data.
  \item $Y_{kd}=\Psi^{-1}(F_{kd}(x_{d}))$ is the d-th dimension of the transferred data.
  \item $Z_{kd}=f_{kd}(x_{d})=\frac{\partial F_{kd}}{\partial x}(x_{d})$ is the density of the marginal distribution.
  \item $\pi_k$ is the weight to the $k$-th copula.
  \end{itemize}
  Its density is given by
  \begin{eqnarray}
  f(X|\pi)=\sum_{k=1}^{K}
   \pi_k\frac{1}{(2\pi)^{n/2}P_k^{1/2}}exp(-\frac{1}{2}Y_{k}^TP_kY_{k})\prod_{d=1}^D\frac{Z_{kd}}{\frac{1}{\sqrt{2\pi}}exp(-\frac{1}{2}(Y_{kd})^2)}
  \end{eqnarray}
  
  The density above is defined conditioned on the cumulative
  probability values and Gaussianized random variables which are both
  determined by the marginal distributions. The marginal distribution
  on each dimension for each component can be estimated via
  nonparametric methods such as kernel smoothing (Bowman 1998 [1]).
  
  %
  
  \section{Basic Properties of GCMM}
  
  A GCMM is defined based on the separation of the mixture of copulas
  and marginal distributions, which may potentially lead to different
  behavior from GMM. To understand the properties of GCMM, its
  likelihood function is studied so that appropriate estimation
  algorithms can be designed. The major properties of GCMM are
  discussed below:
  \begin{itemize}
  \item A GCMM has a bounded likelihood function value on bounded domains and tractable derivatives conditioned on the estimated marginal probability functions. The
  likelihood function is given below:
  \begin{eqnarray}L&=&\sum_{n=1}^{N} ln (\sum_{k=1}^{K} \pi_k
  \frac{1}{(2\pi)^{n/2}P^{1/2}}
  exp(-\frac{1}{2}(Y_{n,k})^TPY_{n,k})\prod_{i=1}^D
  \frac{Z_{n,ki}}{\frac{1}{\sqrt{2\pi}}exp(-\frac{1}{2}(Y_{n,ki})^2) }
  )\end{eqnarray}
  
  We provide the following theorem to demonstrate the features of such
  a likelihood function and the proof is given in the appendix.
  \begin{thm}
  Under suitable conditions, the likelihood function is bounded
  above in bounded region; non-decreasing and negative semi-definite
  w.r.t density $Z_{n,ki}$; may contain both local minimum and local
  maximum w.r.t transformed variables $Y_{n,k}$.
  \end{thm}
  \item The value of its likelihood function is nondecreasing during iterations of Expectation-Maximum algorithms that are applied with GCMM and the algorithms converge globally to local maximums under mild conditions(Wu 1983 [11]). The design and properties of these Expectation-Maximum algorithms are discussed in the next section.
  \item Model selection can be conducted through Akaike information criteria (Fan 2009 [5]) and cluster methods such as k-means or hierarchy clustering can be used to set the initial parameters of each component.
  \end{itemize}
  
  \section{Expectation Maximum Algorithms for GCMM}
  
  \subsection{The Base Case Algorithm}
  
\begin{figure}
  \begin{center}
  \includegraphics[angle=0, width=0.45 \textwidth]{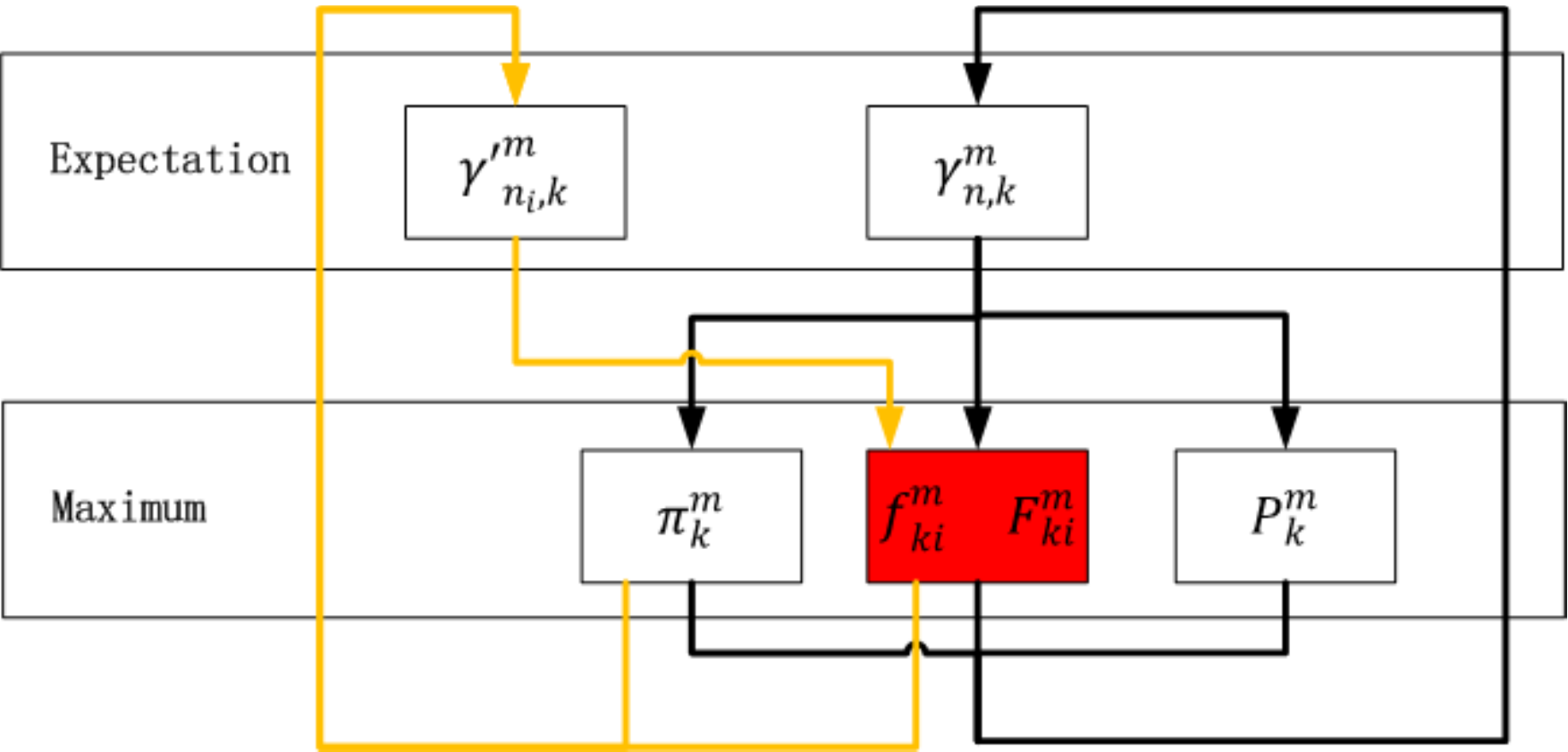} \quad
  \includegraphics[angle=0, width=0.45 \textwidth]{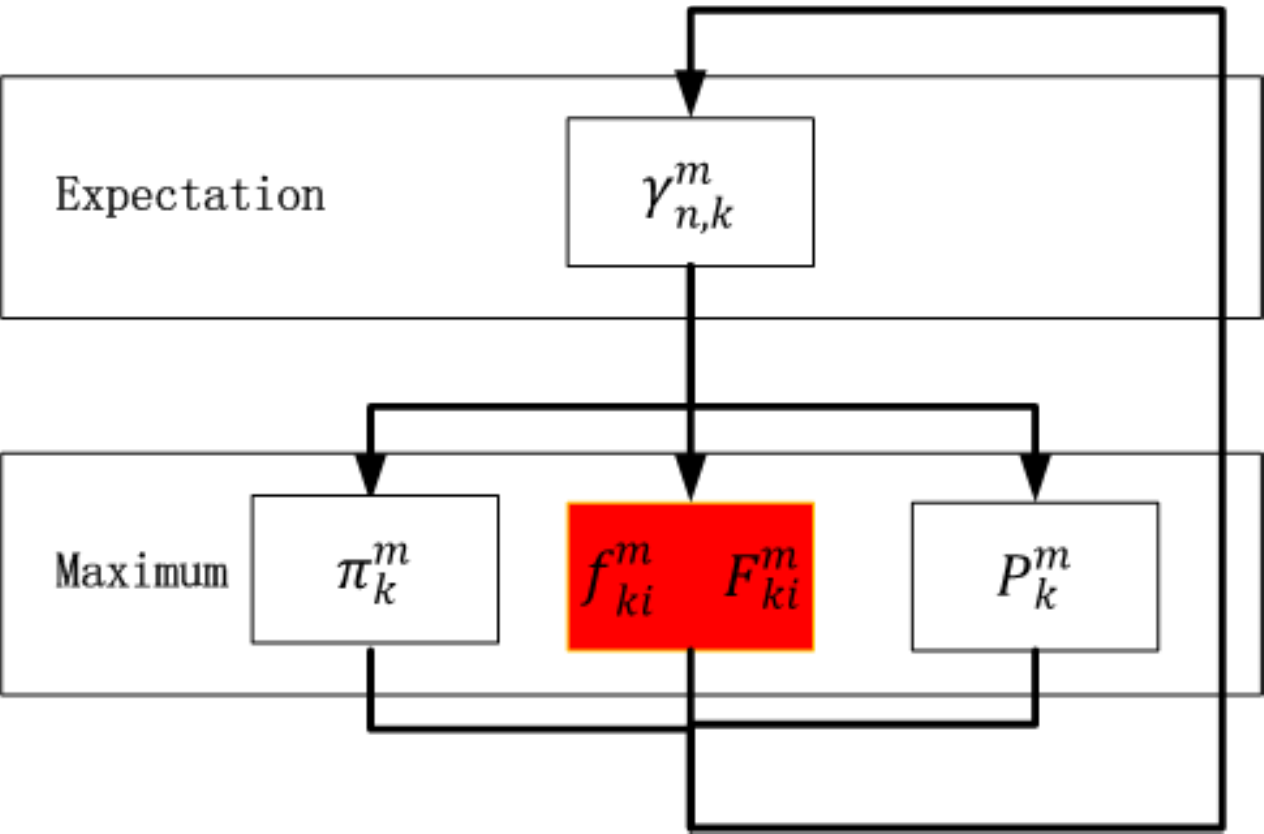}
  \caption{Comparison of GMM and GCMM base case: n: data index; m:
  iteration index; k: copula index; i: dimension index}
  \label{fig:5-GMMGCMM}
  \end{center}
  \end{figure}

  The algorithm updates the mixture of copulas and the marginal
  distributions separately. Essentially when estimating GMMs, the
  weights $\pi^m_{k}$ \& correlation matrixes of components $P^m_k$
  and the sufficient statistics (mean $\mu^m_{ki}$ and standard
  deviation $\sigma^m_{ki}$) of the marginal normal distributions are
  updated (Dempster 1977 [3]) based on the posterior probability
  $\gamma^m_{nk}$. In GCMMs, the sufficient statistics of marginal
  normal distributions are replaced with non-parametric estimators to
  the marginal pdf $f^m_{ki}$ and cdf $F^m_{ki}$ to improve flexility,
  see the red boxes in Figure \ref{fig:5-GMMGCMM}.

  The major challenge of algorithm design lies in how the marginal
  distributions should be updated considering the posterior
  probability. An updating formula is developed and given by the
  following theorem:
  
  \begin{thm}
  In the GCMM base case, the updating of the marginal distributions
  follows the following formula  with necessary normalizations:
  $$F'_{ki}(c)=\sum_n \gamma_{nk} 1_{x_{ni} \leq c}$$
  \end{thm}
  
  Based on the theorem, the algorithm is further developed below:
  
  \begin{itemize}
  \item Expectation Step:
  \begin{eqnarray}
  D^m_{nk}&=&\prod_{i=1}^D\frac{Z^m_{n,ki}}{\frac{1}{\sqrt{2\pi}}\exp(\frac{1}{2}(Y^m_{n,ki})^2)}
  \end{eqnarray}
  \begin{eqnarray}
  r^m_{nk}=\frac{\pi_k
  \frac{D^m_{n,k}}{|P^m_k|^{1/2}}exp(-\frac{1}{2}(Y^m_{nk})^TP_k^{m,-1}Y^m_{nk})}{\sum_{j=1}^K
  \pi_j\frac{D^m_{n,j}}{|P^m_j|^{1/2}}exp(-\frac{1}{2}(Y^m_{nj})^TP_j^{m,-1}Y^m_{nj})}
  \end{eqnarray}
  \item Maximum Step:
  \begin{eqnarray}
  \pi_k^m&=&\frac{\sum_{n=1}^N r^m_{nk}}{N}
   \end{eqnarray}
  \begin{eqnarray}
  P_k^m&=&\frac{\sum_{n=1}^{N}r^m_{nk}Y^m_{nk}(Y^m_{nk})^T}{\sum_{n=1}^{N}r^m_{nk}}
  \end{eqnarray}
  \begin{eqnarray}
  F_{ki}^m(y)=\frac{\sum_n r_{nk}^m 1_{x_{ni} \leq y}}{\sum_n
  r_{nk}^m}
  \end{eqnarray}
  $\forall k$-th copula, $i$-th dimension
  \end{itemize}
  
  The issue here is that heavy tail phenomenons may be categorized
  into two classes: the heavy tails in the marginal distribution and
  the heavy tails in the dependence structure. GCMM separates the
  estimation for them and control the number of clusters purely based
  on the complexity of heavy tails in dependence structure (the
  latter). In this manner, the number of clusters could be further
  reduced and the mixture of copulas are robust towards heavy tails on
  the marginal distributions (the former).
  \subsection{With Unsynchronized Data}
  
\begin{figure}
  \begin{center}
  \includegraphics[angle=0, width=0.5 \textwidth]{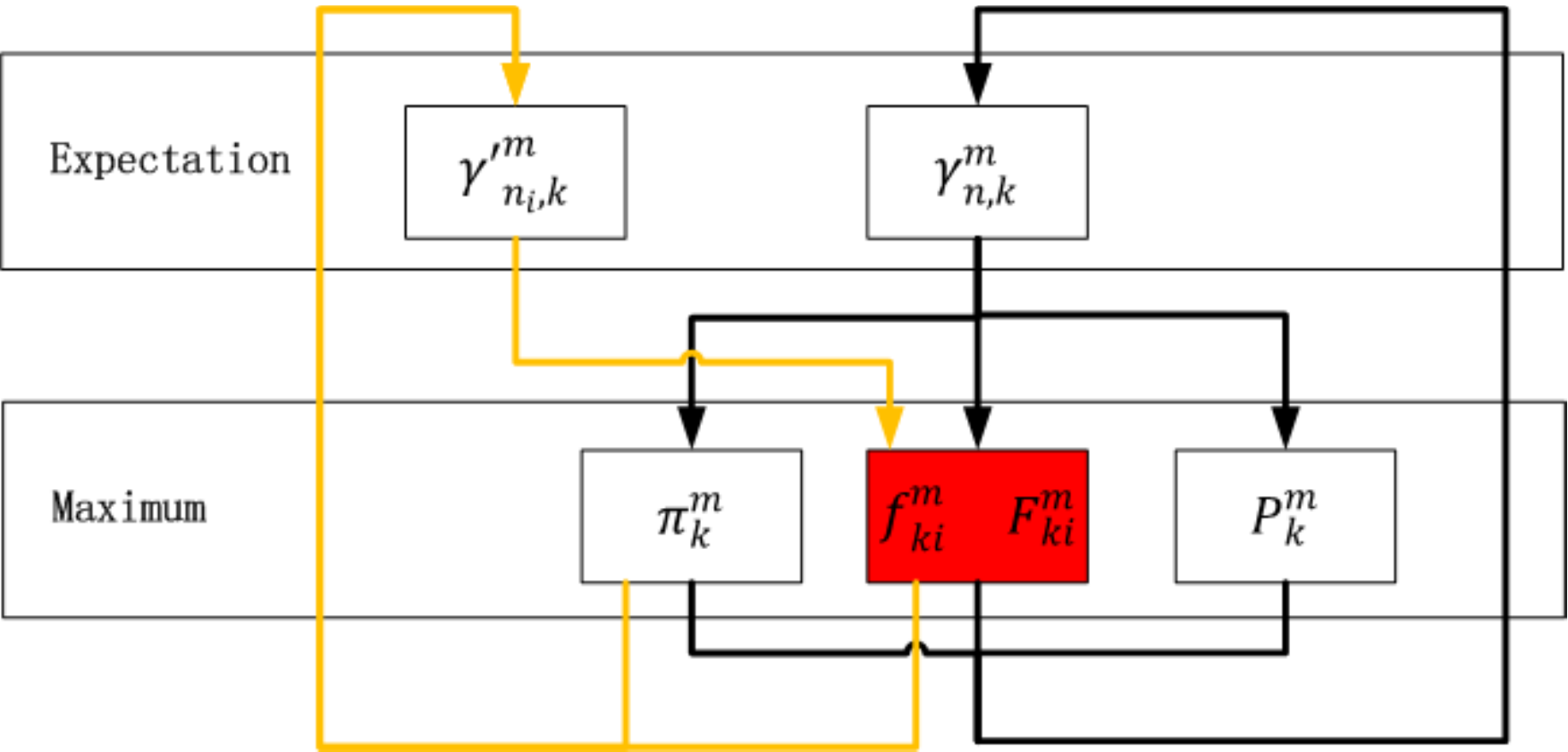}
  \caption{Comparison of GCMM base case and GCMM with unsynchronized
  data: n: data index; m: iteration index; k: copula index; i:
  dimension} \label{fig:5-GCMMbvsa}
  \end{center}
  \end{figure}

  GCMMs with unsynchronized data are developed based on the rationale
  that unsynchronized data in each dimension can be used to update the
  marginal distribution, given the estimation of marginal distribution
  is separated from the mixture of copulas. An additional posterior
  probability $\gamma'^m_{n_i,k}$ is introduced to represent the
  probability of $n_i$-th unsynchronized data on the $i$-th dimension
  belonging to the $k$-th component. An additional loop is then
  inserted into the Expectation Maximum algorithm for GCMM base case
  which further updates $\gamma'^m_{n_i,k}$ based on new information,
  see the orange loop in Figure \ref{fig:5-GCMMbvsa}.
  
  The major challenge of algorithm design lies in how the marginal
  distributions should be further updated given unsynchronized data
  and the existing nonparametric estimator. An updating formula is
  developed and given by the following theorem:
  
  \begin{thm}
  In the GCMM with unsynchronized data, the updating formula of
  marginal distribution follows by the following formula with
  necessary normalizations:
  $$r'_{n_i,k}=\frac{\pi_k f_{ki}(x_{n_i})}{\sum_{k=1}^K \pi_k
  f_{ki}(x_{n_i})}$$
  $$F'_{ki}(c)=\sum_n \gamma_{nk} 1_{x_{ni} \leq c}+\sum_{n_i}
  \gamma'_{n_i,k} 1_{x_{n_i} \leq c}$$
  \end{thm}
  Based on the theorem, the algorithm is further developed below
  (similar parts as the base case are ignored to save space):
  
  \begin{itemize}
  \item In Expectation step:
  \begin{itemize}
  \item update $r_{nk}^m$ for synchronized data;
  \item update $r^{'m}_{n_i,k}$ for un-synchronized data using the following Bayes
  formula:
  \begin{eqnarray}
  r^{'m}_{n_i,k}=\frac{\pi_k^m f_{ki}^m(x_{n_i})}{\sum_{k=1}^K \pi_k^m
  f_{ki}^m(x_{n_i})}
  \end{eqnarray}
  \end{itemize}
  \item In each iteration, update the marginal cdfs $F_n$ according to $r_{nk}$ and $r'_{n'k}$. $\forall$ k-th copula, i-th
  dimension:
  \begin{eqnarray}
  F^m_{ki}(y)=\frac{\sum_n r^m_{nk} 1_{x_{ni}\leq y}+\sum_{n_i}
  r^{'m}_{n_i,k} 1_{x_{n_i}\leq y}}{\sum_n r^m_{nk}+\sum_{n_i}
  r^{'m}_{n_i,k}}
  \end{eqnarray}
  \end{itemize}
  
  The philosophical issue here is whether synchronized data truly
  represent the joint distribution adequately and whether the
  unsynchronized data may add to our understanding of it. To bring
  unsynchronized data into the whole Expectation Maximum algorithm
  enlarges the information set of the probability space
  ($\Omega,\mathcal{F},P$) so that deeper elaboration of the data is
  possible (Cinlar 2011 [2]). This is a significant improvement from
  GMM beyond the flexibility applied to the marginal distribution.
  
  \section{Experiment}
  
  \subsection{Simulation Test}
  
  In this section, two-dimensional data are simulated based on a
  three-copula GCMM and the distribution of the data is given in
  Figure \ref{fig:5-cctftl}. Then the two Expectation Maximum
  algorithms are utilized to estimate the model and Akaike information
  critera is used to select the number of clusters. It is found that
  GMM needs five clusters to explain the data well while GCMM needs
  three. We further aggregate the data in the three dimensions to see
  the fitting for their sum: additional data are simulated with the
  estimated GMM and GCMM and their sum is compared with that for the
  calibration data. Two sample KS test demonstrates that the simulated
  data based on GCMM captures the distribution of the calibration data
  set.
  \begin{itemize}
  \item GCMM achieves better fitting with fewer clusters.
  \begin{figure}[htb]
  \begin{center}
   {\includegraphics[angle =0 , width=0.3
  \textwidth]{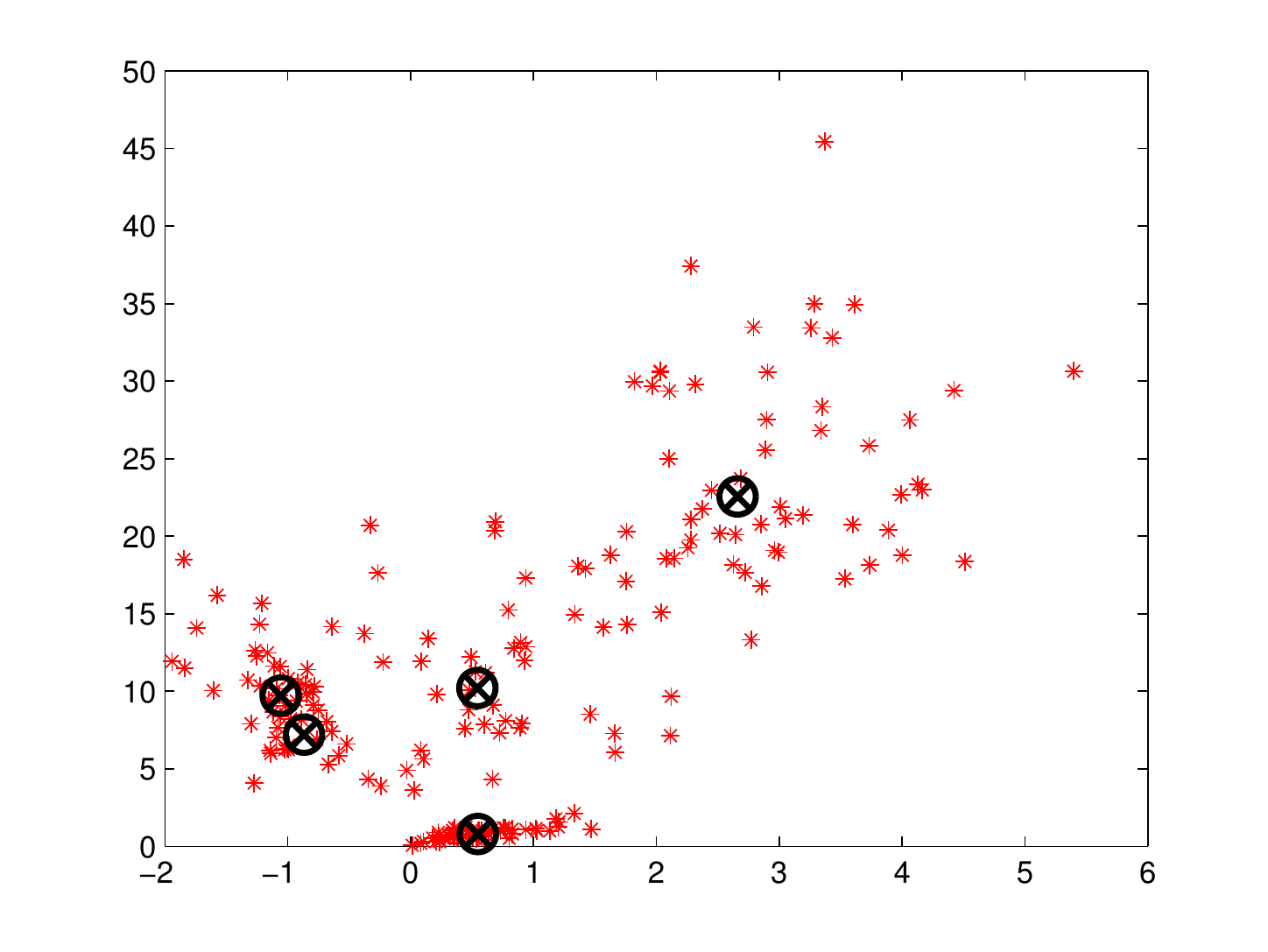}}
  {\includegraphics[angle =0 , width=0.3
  \textwidth]{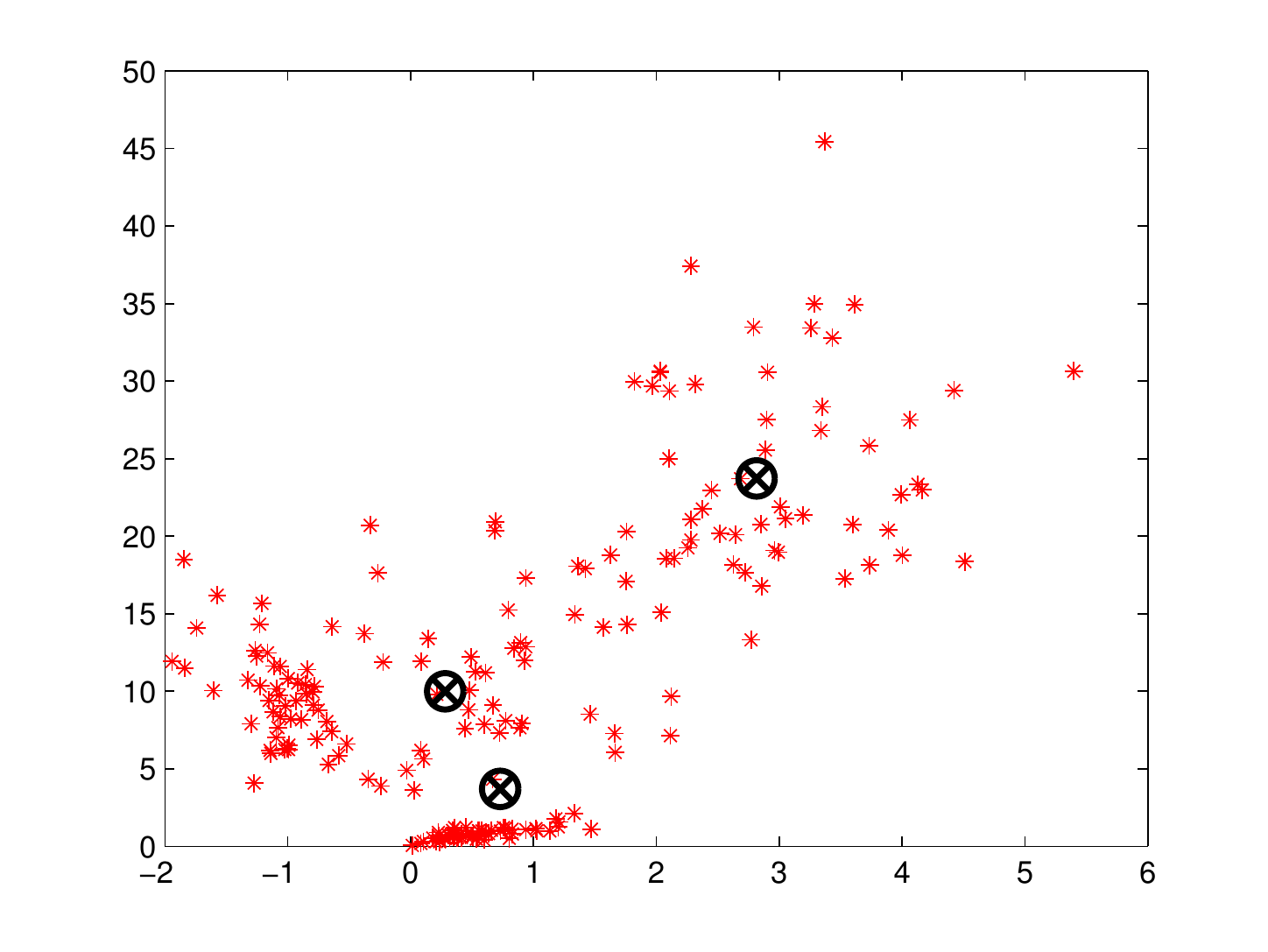}}\quad
  {\includegraphics[angle=0 , width=0.3
  \textwidth]{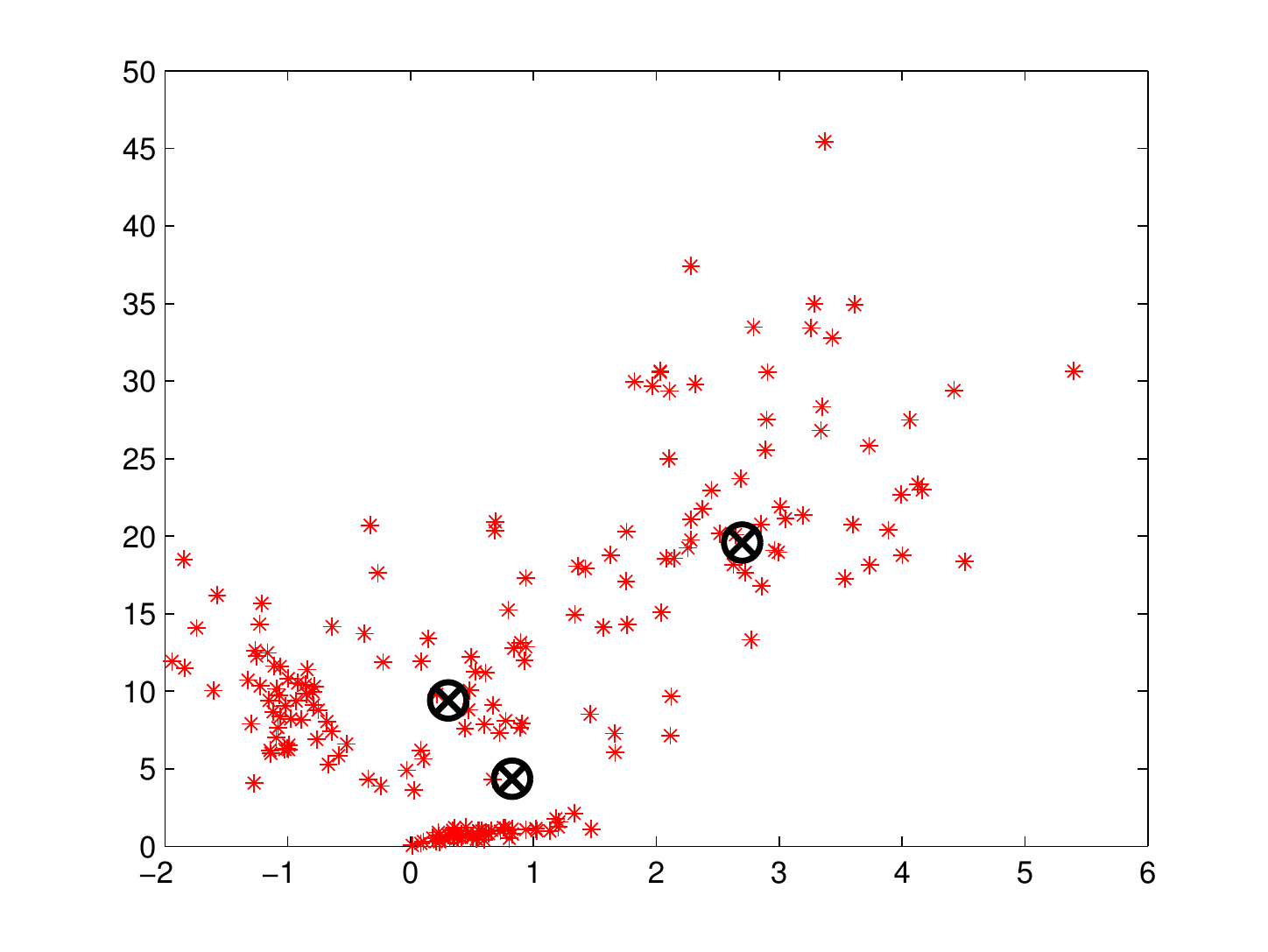}}
  \caption{Clusters for GMM v.s. Clusters for
  GCMM}\label{fig:5-cctftl}
  \end{center}
  \end{figure}
  \item The p-values of two sample KS test for the sum of two random variables are compared in
  Table \ref{tab:simutab}, which suggests that the GCMM fits the
  distribution of sum better than the GMM given the same number of
  clusters.
  \begin{table}
  \caption{p-values of two-sample KS test compared with the simulated
  distribution}  \centering \label{tab:simutab}
  \begin{tabular}{|c|c|c|}
  \hline
  GMM  & Base Case & Extra-Data\\
  \hline
  0.00\textcolor[rgb]{0.75,0.75,0.75}{02} & 0.13\textcolor[rgb]{0.75,0.75,0.75}{04}&0.10\textcolor[rgb]{0.75,0.75,0.75}{03}\\
  \hline
  \end{tabular}
  \end{table}
  \end{itemize}

  \subsection{Test on Empirical Data}
  
  A real data set from the transportation system using the travel time
  of individual drivers in New Jersey which is captured from GPS
  devices is employed for model testing. On each transportation link
  (a road segment) there are many travel time observations, and by
  matching the departure time of the current link and the arrival time
  of the immediate downstream link, such data can be synchronized to
  construct the vector for running GMM. However, not all data on each
  link can be synchronized because the arrival times of drivers are
  random and sparse in time. The ultimate goal is to aggregate such
  link level data for estimating the distribution of the travel time
  over a path consisting of a few consecutive links. The same
  procedure is used as the simulation test in the previous section
  except the calibration data set is real. The results are summarized
  below, to save space the three-dimensional clusters are omitted:
  \begin{itemize}
  \item The comparison to the empirical path travel time distribution is shown in Table \ref{tab:5-ksreal} for a three segment
  path. Akaike information criteria indicates both the GMM and the
  GCMM needs three clusters to describe the data well and p-values of
  the KS tests for GCMM are noticeably larger.
  \begin{table}
  \caption{p-values of two-sample KS test compared with the empirical
  distribution}  \label{tab:5-ksreal}  \centering
  \begin{tabular}{|c|c|c|}
  \hline
  GMM  & Base Case & Extra-Data\\
  \hline
  0.05\textcolor[rgb]{0.75,0.75,0.75}{18} & 0.96\textcolor[rgb]{0.75,0.75,0.75}{46}&0.11\textcolor[rgb]{0.75,0.75,0.75}{57}\\
  \hline
  \end{tabular}
  \end{table}
  \item Estimated distributions are compared in Figure \ref{fig:5-cctftlreal}, GCMM with
  unsynchronized data captures heavier tails as there are some higher
  values in the unsynchronized data. The heavier tail is caused by
  differences in marginal distributions due to new information in the
  synchronized data, but not by material changes of the mixture of
  copulas.
  \begin{figure}[htb]
  \begin{center}
   {\includegraphics[angle =0 , width=0.4\textwidth]{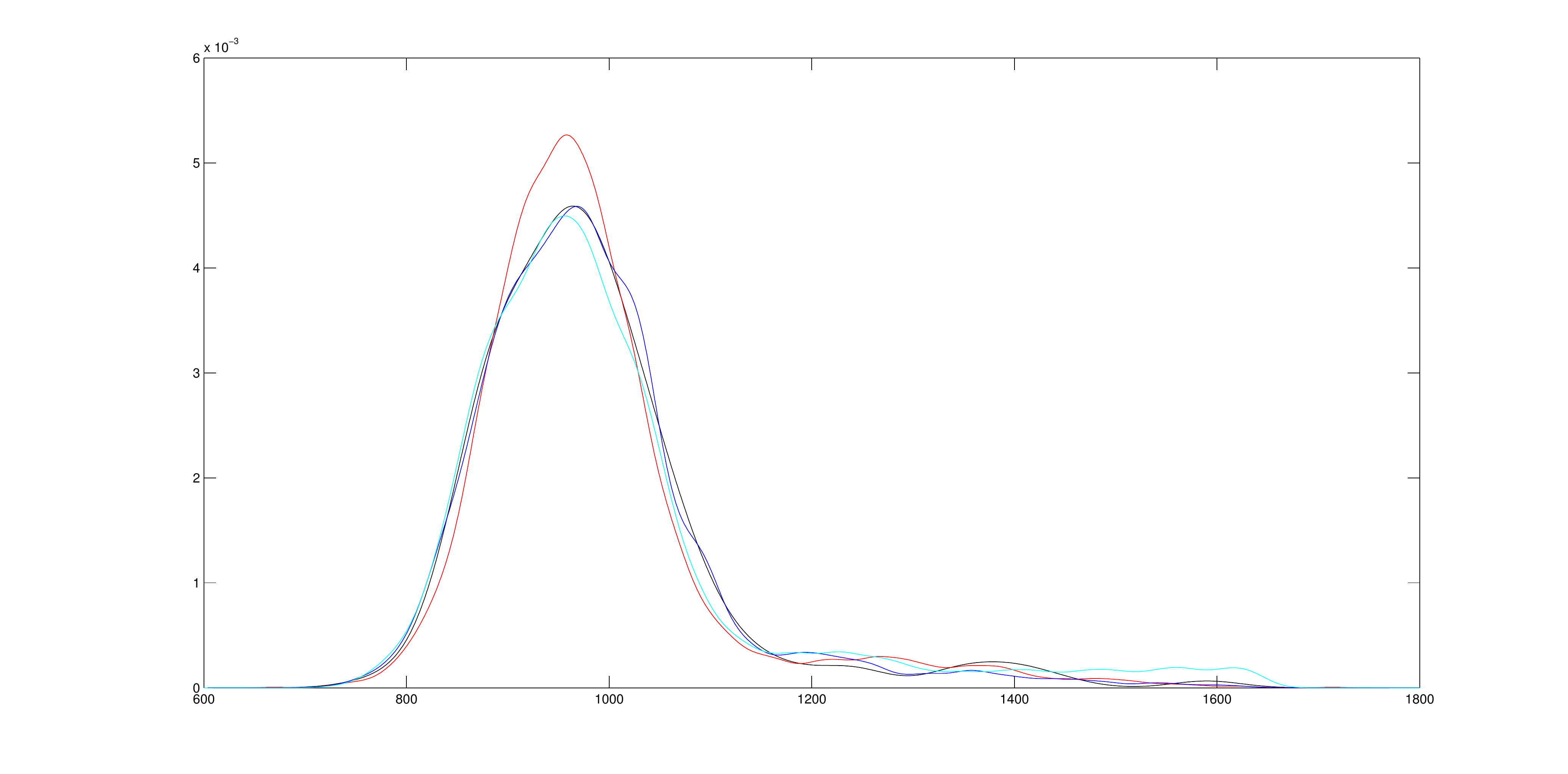}}
   {\includegraphics[angle=0 , width=0.28
  \textwidth]{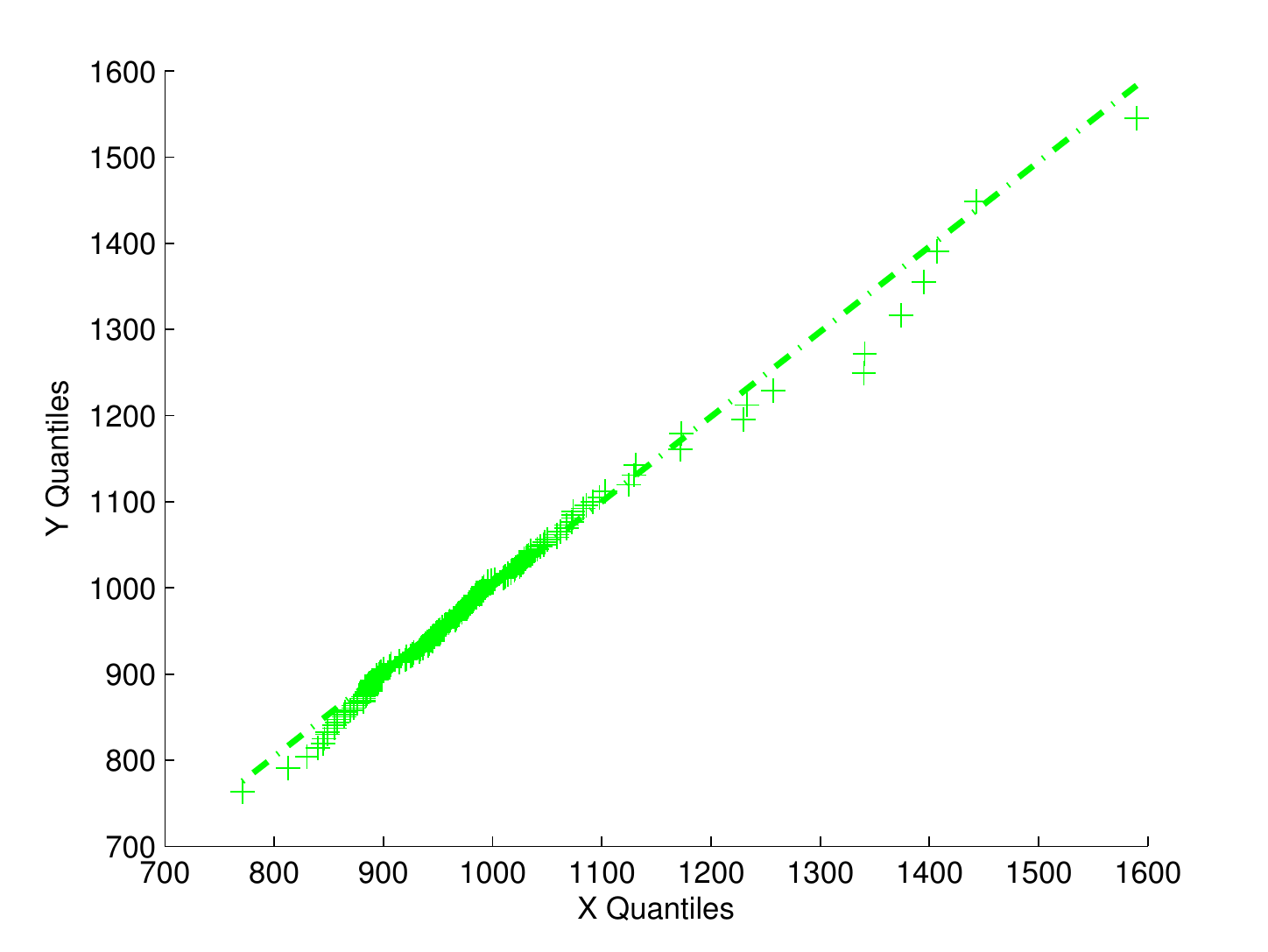}}
  {\includegraphics[angle=0 , width=0.28
  \textwidth]{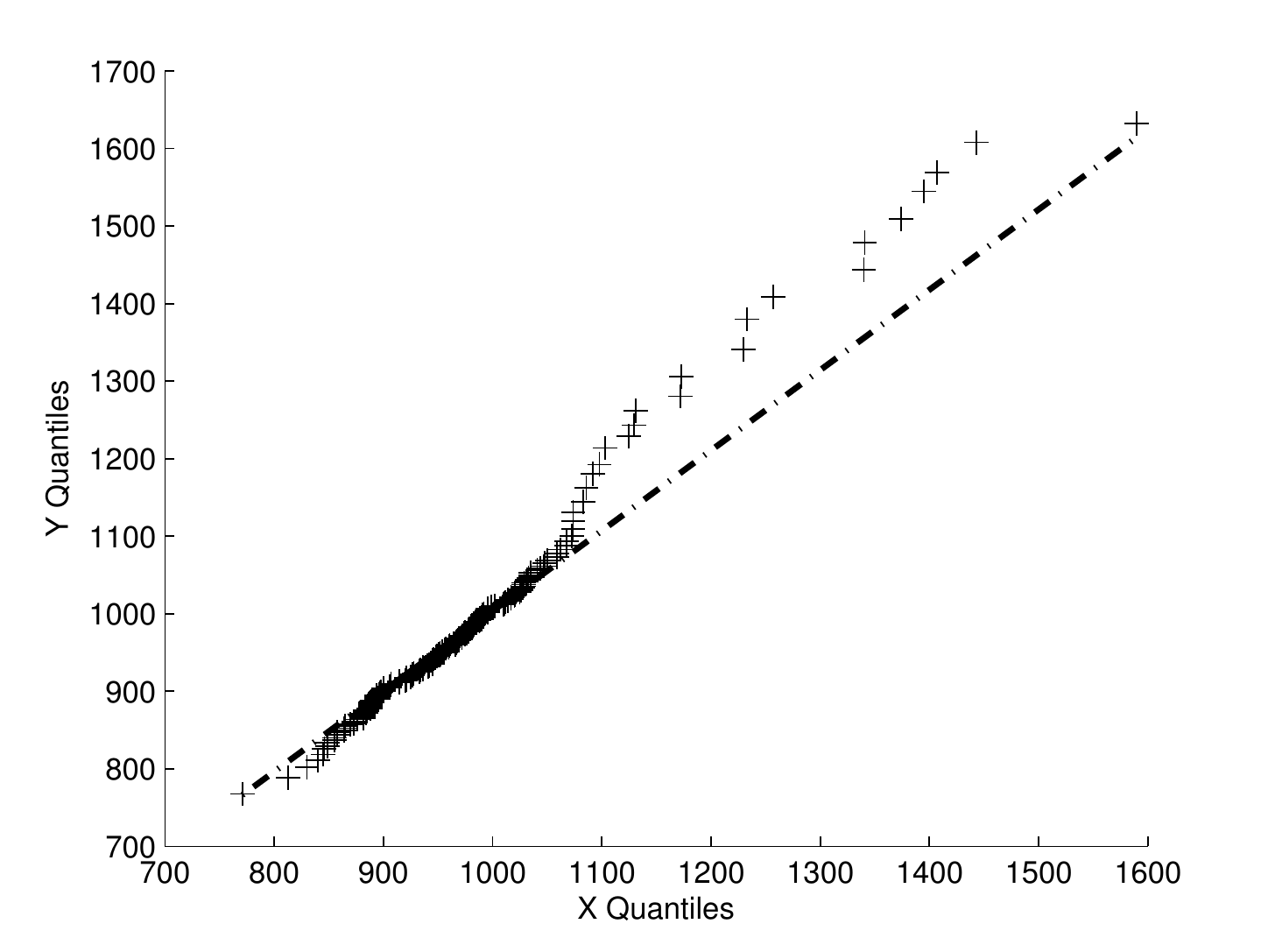}}
  \caption{Comparison of pdf (Red: GMM; Blue: GCMM base case; Cyan:
  GCMM with unsynchronized data; Black: Empirical); Green QQplot
  Empirical(x) v.s. GCMM base case(y); Black QQplot Empirical(x) v.s.
  GCMM with unsynchronized data(y)}
  \label{fig:5-cctftlreal}
  \end{center}
  \end{figure}
  \end{itemize}
  
  \section{Conclusion}
  
  In this paper, Gaussian copula mixture models (GCMMs) are developed
  to estimate the joint distribution of a group of random variables
  and further estimate the distribution of their sum.  The Expectation
  Maximum algorithm is extended to estimate the GCMM models. Overall,
  GCMMs first add more flexibility to fit heavy tails on marginal
  distributions while remaining relatively robust against it; GCMMs
  further incorporate unsynchronized data into estimation, both of
  which improve the approximation to the complex dependence structures
  given limited number of components. In the future, the empirical
  properties of this new category of models on specific data sets can
  be studied further.
  
  \section{Proofs}
  
  \subsection{Proof to Theorem 1}
  
  Proof  Consider maximizing the following function
  \begin{eqnarray}L=\sum_{n=1}^{N} ln (\sum_{k=1}^{K} \pi_k
  \frac{1}{(2\pi)^{n/2}P^{1/2}}
  exp(-\frac{1}{2}(Y_{n,k})^TPY_{n,k})\prod_{i=1}^D
  \frac{Z_{n,ki}}{\frac{1}{\sqrt{2\pi}}exp(-\frac{1}{2}(Y_{n,ki})^2) }
  )\end{eqnarray}
   with the constraints:
  \begin{center}
  $Z_{n,k}\succeq 0$, $Z_{n,k}\preceq C$, $Y_{n,k}\succeq0$ and
  $Y_{n,k}\prec 1$
  \end{center}
   If it is changed into a minimization problem by multiplying the objective by -1, the full Lagrange objective function will be:
  \begin{eqnarray*}
  \widehat{L}&=&-\sum_{n=1}^{N} ln (\sum_{k=1}^{K} \pi_k
  \frac{1}{(2\pi)^{n/2}P^{1/2}}
  exp(-\frac{1}{2}(Y_{n,k})^TPY_{n,k})\prod_{i=1}^D
  \frac{Z_{n,ki}}{\frac{1}{\sqrt{2\pi}}exp(-\frac{1}{2}(Y_{n,ki})^2) }
  )\\
  &&+\sum_{n}\sum_{k}\alpha_{n,k}^T(-Y_{n,k})+\sum_{n}\sum_{k}\beta_{n,k}^T(Y_{n,k}-1)+\sum_{n}\sum_{k}\gamma_{n,k}^T(-Z_{n,k})\\
  &&+\sum_{n}\sum_{k}\theta_{n,k}^T(Z_{n,k}-C)
  \end{eqnarray*}
  with \begin{center} $\alpha_{n,k}\succeq 0$,
   $\gamma_{n,k}\succeq 0$,
   $\beta_{n,k}\succeq 0$,
  $\theta_{n,k}\succeq 0$
  \end{center}
  and
  \begin{center} $-Z_{n,k}\preceq 0$, $Z_{n,k}-C \preceq 0$, $-Y_{n,k}\preceq0$ and
  $Y_{n,k}-1 \prec 0$
  \end{center}
  
   Then
  \begin{eqnarray*}
  \frac{\partial L}{\partial Z_{n,kj}}&=& \frac{1}{S_{n}} \pi_k \frac{1}{(2\pi)^{n/2}P^{1/2}}exp(-\frac{1}{2}Y_{n,k}^TPY_{n,k})\prod_{i\neq j}\frac{Z_{n,ki}}{\frac{1}{\sqrt{2\pi}}exp(-\frac{1}{2}Y_{n,ki}^2)}\\
  &\geq&0\\
  \end{eqnarray*}
  Where $S_{n}=\sum_{k=1}^{K} \pi_k
  \frac{1}{(2\pi)^{n/2}P^{1/2}}exp(-\frac{1}{2}Y_{n,k}^TPY_{n,k})\prod_{i=1}^D\frac{Z_{n,ki}}{\frac{1}{\sqrt{2\pi}}exp(-\frac{1}{2}Y^2)}$
  \begin{eqnarray*}
  \frac{\partial L^2}{\partial Z_{n,kj}^2}&=&
  \frac{1}{S^2_{n}}(0-P_{n,kj}^2)\\
  &\leq&0\\
  \end{eqnarray*}
  where $P_{n,kj}= \pi_k
  \frac{1}{(2\pi)^{n/2}P^{1/2}}exp(-\frac{1}{2}Y_{n,k}^TPY_{n,k})\prod_{i\neq
  j}\frac{Z_{n,ki}}{\frac{1}{\sqrt{2\pi}}exp(-\frac{1}{2}Y_{n,ki}^2)}$
  And
  \begin{eqnarray*}
  \frac{\partial \widehat{L}}{\partial Z_{n,kj}}&=& -\frac{1}{S_{n}} \pi_k \frac{1}{(2\pi)^{n/2}P^{1/2}}exp(-\frac{1}{2}Y_{n,k}^TPY_{n,k})\prod_{i\neq j}\frac{Z_{n,ki}}{\frac{1}{\sqrt{2\pi}}exp(-\frac{1}{2}Y_{n,ki}^2)}-\gamma_{n,kj}+\theta_{n,kj}\\
  \end{eqnarray*}
  $$\gamma_{n,k}^TZ_{n,k}= 0$$,
  $$\theta_{n,k}^T(Z_{n,k}-C)=0$$
  $$\frac{\partial \widehat{L}^2}{\partial Z^2_{n,kj}}\geq 0$$
  By taking $\frac{\partial \widehat{L}}{\partial Z_{n,kj}}=0$, there
  should be the following relationship:
  \begin{eqnarray}
  Z_{n_ki}=\frac{S_n(-\gamma_{n,kj}+\theta_{n,kj})}{D_{n,k}\frac{1}{2\pi}exp(-\frac{1}{2}Y^2_{n,ki})}
  \end{eqnarray}
  \begin{eqnarray}
  \gamma_{n,ki}\frac{S_n(-\gamma_{n,kj}+\theta_{n,kj})}{D_{n,k}\frac{1}{2\pi}exp(-\frac{1}{2}Y^2_{n,ki})}=0
  \end{eqnarray}
  \begin{eqnarray}
  \theta_{n,ki}(\frac{S_n(-\gamma_{n,kj}+\theta_{n,kj})}{D_{n,k}\frac{1}{2\pi}exp(-\frac{1}{2}Y^2_{n,ki})}-C_i)=0
  \end{eqnarray}
  The objective $\widehat{L}$ may be minimized in the inner area, that
  is: (1)$\gamma_{n,kj}=0$ and $\theta_{n,kj}\neq 0$ the solution is
  denoted as $Z^{b1}_{n,k}$; (2) $\gamma_{n,kj} \neq 0$ and
  $\theta_{n,kj}=0$, the solution is denoted as $Z^{b2}_{n,k}$; (3)
  $\gamma_{n,kj}=0$ and $\theta_{n,kj}= 0$, the solution is denoted as
  $Z^{c}_{n,k}$.
  
  For $Y_{n,k}$, the following analysis is conducted:
  \begin{eqnarray*}
  \frac{\partial L}{\partial Y_{n,k}}&=& \frac{1}{S_{n}}
   B_{n,k}D_{n,k}(-P+I)Y_{n,k}\\
  \end{eqnarray*}
  where $B_{n,k}=\pi_k
  \frac{1}{(2\pi)^{n/2}P^{1/2}}exp(-\frac{1}{2}Y_{n,k}^TPY_{n,k})$ and
  $D_{n,k}=\prod_{i=1}^D\frac{Z_{n,ki}}{\frac{1}{\sqrt{2\pi}}exp(-\frac{1}{2}Y_{n,ki}^2)}$
  as defined in the previous section
  
  \begin{eqnarray*}
  \frac{\partial L^2}{\partial Y^2_{n,k}}&=& \frac{1}{S^2_{n}}
   (B_{n,k}D_{n,k}(-P+I)Y_{n,k}Y_{n,k}^T(-P+I)+B_{n,k}D_{n,k}(-P+I))S_n\\
   &&-B_{n,k}D_{n,k}(-P+I)Y_{n,k}Y_{n,k}^T(-P+I)B_{n,k}D_{n,k}\\
   &=&\frac{1}{S^2_{n}}(B_{n,k}D_{n,k}(-P+I)(Y_{n,k}Y_{n,k}^TS_n+S_n-Y_{n,k}Y_{n,k}^TB_{n,k}D_{n,k})(-P+I)\\
   &=&\frac{1}{S^2_{n}}(B_{n,k}D_{n,k}(Y_{n,k}Y_{n,k}^TS_n+S_n-B_{n,k}D_{n,k}Y_{n,k}Y_{n,k}^T)(-P+I)(-P+I)\\
  \end{eqnarray*}
  Notice here $(-P+I)$ is diagonalizable since $P$ is the covariance
  matrix of two normally distributed random vectors. $(-P+I)(-P+I)$ is
  then positive and semi-definite. Define
  $$\Lambda_{n,k}=Y_{n,k}Y_{n,k}^TS_n+S_n-B_{n,k}D_{n,k}Y_{n,k}Y_{n,k}^T$$ Since
  $\frac{1}{S^2_{n}}B_{n,k}D_{n,k}$ is positive, $\Lambda_{n,k}$ will
  determine the properties of the function with respect to $Y_{n,k}$.
  $$\frac{1}{S_{n}} B_{n,k}D_{n,k}(-P+I)Y_{n,k}-\alpha_{n,k}+\beta_{n,k}=0$$
  $$\alpha_{n,k}^TY_{n,k}=0$$
  $$\beta_{n,k}^T(Y_{n,k}-1)=0$$
  \begin{center} if $\Lambda_{n,k}\geq0$, $\frac{\partial L^2}{\partial
  Y^2_{n,k}}\succeq 0$ and $\frac{\partial \widehat{L}^2}{\partial
  Y^2_{n,k}}\preceq 0$
  
  if $\Lambda_{n,k}<0$, $\frac{\partial L^2}{\partial Y^2_{n,k}}\prec
  0$ and $\frac{\partial \widehat{L}^2}{\partial Y^2_{n,k}} \succ 0$
  \end{center}
   Then
  \begin{eqnarray}
  Y_{n,k}&=&\frac{S_{n}}{B_{n,k}D_{n,k}}(-P+I)^{-1}(\alpha_{n,k}-\beta_{n,k})
  \end{eqnarray}
  \begin{eqnarray}
  \frac{S_{n}}{B_{n,k}D_{n,k}}\alpha_{n,k}^T(-P+I)^{-1}(\alpha_{n,k}-\beta_{n,k})=0
  \end{eqnarray}
  \begin{eqnarray}
  \beta_{n,k}^T(\frac{S_{n}}{B_{n,k}D_{n,k}}(-P+I)^{-1}(\alpha_{n,k}-\beta_{n,k})-1)=0
  \end{eqnarray}
  
  Then $Y_{n,k}$ can be solved using the equations above. Furthermore,
  extreme values of the $Y_{n,k}$ are considered as follows:
  
  If $\Lambda_{n,k}\geq 0$ then the data point $x_n$ is classified as
  Type 1 for k-th copula. The objective$\widehat{L}$ is always
  minimized on the boundary. That is: (1) $\alpha_{n,k}=0$ and
  $\beta_{n,k}\neq 0$, the solution is denoted as $Y^{1b1}_{n,k}$;
  (2)$\alpha_{n,k} \neq 0$ and $\beta_{n,k}=0$ the solution is denoted
  as $Y^{1b2}_{n,k}$.
  
   If $\Lambda_{n,k}<0$, then the data point $x_n$ is classified as Type 2 for k-th copula. The objective
  $\widehat{L}$ may be minimized in the inner area. That is:
  (1)$\alpha_{n,k}=0$ and $\beta_{n,k}\neq 0$, the solution is denoted
  as $Y^{2b1}_{n,k}$; (2) $\alpha_{n,k} \neq 0$ and $\beta_{n,k}=0$,
  the solution is denoted as $Y^{2b2}_{n,k}$; (3) $\alpha_{n,k}=0$ and
  $\beta_{n,k}= 0$, the solution is denoted as $Y^{2c}_{n,k}$.
  
   In all cases, the value of the likelihood function is bounded above by a value determined by these finite extreme values in $Y_{n,k}$ and $Z_{n,k}$.
  Q.E.D.

  \subsection{Proof to Theorem 2}
  
  Denote $x_n$ as the observed synchronized data vector,$z$ are the
  complete data. Recall in the Expectation step we calculate the
  posterior probability $\gamma_{nk}$ for $n$-th data vector belong to
  $k$-th cluster such that the incomplete data likelihood function
  below is expressed explicitly.
  $Q(\pi',P',F'|\pi,P,F)=E(logf(z)|x_n,\pi,P,F )$
  
  In the Maximum step we calculate $[\pi',P',F']=argmax_{\pi',P',F'}
  Q(\pi',P',F'|\pi,P,F)$ to obtain new parameters based on such
  $\gamma_{nk}$.
  
  In this process, the poster distribution $\gamma_{nk}$ is
  $\gamma_{nk}=p_k(x_n|\pi,P,F)$
  
  So the natural estimator for the marginal distribution for the
  $i$-th dimension of the $k$-th component is its histogram
  conditioned on the current weights: $F'_{ki}(c)=p_{ki}(x_{ni} \leq
  c|\pi,P,F)=\sum_n p_{k}(x_{ni}|\pi,P,F)1_{x_{ni} \leq c}=\sum_n
  \gamma_{nk} 1_{x_{ni} \leq c}$
  
  Further normalization is used to maintain the properties of a cdf
  and other univariate non-parametric estimator can be used. Q.E.D
  
  \subsection{Proof to Theorem 3}
  
  Denote $x_n$ as the observed synchronized data vector, $x_{n_i}$ as
  $n_i$-th observed unsynchronized data on the i-th dimension and $z$
  as the complete data Recall in the Expectation step of the
  likelihood function is to calculate the posterior probability
  $\gamma_{nk}$ for $n$-th data vector belong to $k$-th cluster such
  that the incomplete data likelihood function below is expressed
  explicitly. $Q(\pi',P',F'|\pi,P,F)=E(logf(z)|x_{n},x_{n_i},\pi,P,F
  )$
  
  Moreover, we also calculate the posterior probability
  $\gamma'_{n_i,k}$ for $x_{n_i}$ (the $n_i$-th unsynchronized
  observation on the i-th dimension) to belong to $k$-th cluster based
  on $F_{ki}(c)$.
  
  In the Maximum step we calculate $[\pi',P',F']=argmax_{\pi',P',F'}
  Q(\pi',P',F'|\pi,P,F)$ to obtain new parameters based on such
  $\gamma_{nk}$ and $\gamma'_{n_i,k}$.
  
  The poster distribution $\gamma_{nk}$ is
  $\gamma_{nk}=p_k(x_n|\pi,P,F)$
  
  The poster distribution $\gamma'_{n_i,k}$ is
  $\gamma'_{n_i,k}=p_k(x_{n_i}|\pi,P,F)=\frac{p_k(x_{n_i}|\pi,P,F,K=k)P(K=k)}{\sum_k
  p_k(x_{n_i}|\pi,P,F,K=k)P(K=k)}=\frac{\pi_k f_{ki}(x_{n_i})}{\sum_k
  \pi_k f_{ki}(x_{n_i})}$
  
  So the natural estimator for the marginal distribution for the
  $i$-th dimension of the $k$-th component is its histogram
  conditioned on the current weights for all data on that dimension.
  $F'_{ki}(c)=p_{k}(x_{ni} \leq c|\pi,P,F)+p_{k}(x_{n_i} \leq
  c|\pi,P,F)=\sum_n p_{k}(x_{ni}|\pi,P,F)1_{x_{ni} \leq c}+\sum_{n_i}
  p_{k}(x_{n_i}|\pi,P,F)1_{x_{n_i} \leq c}=\sum_n \gamma_{nk}
  1_{x_{ni} \leq c}+\sum_{n_i} \gamma'_{n_i,k} 1_{x_{n_i} \leq c}$
  
  Further normalization is used to maintain the properties of a cdf
  and other univariate non-parametric estimators can be used. Q.E.D
  
  
  \section{Acknowledgement}

  \section{Reference}

  \small{

  [1] Bowman, A., Hall, P. \& Prvan, T. (1998), `Bandwidth selection
  for the smoothing of distribution functions', Biometrika 85(4), 799.
  
  [2] Erhan Cinlar, Probability and stochastics, volume 261. Springer,
  2011.
  
  [3] Arthur P Dempster, Nan M Laird, Donald B Rubin, et al. Maximum
  likelihood from incomplete data via the em algorithm. Journal of the
  Royal statistical Society, 39(1):1-38, 1977
  
  [4] Dong Hwan Oh and Andrew J. Patton, Modelling Dependence in High
  Dimensions with Factor Copulas, Duke University 31 May 2011
  
  [5] Jianqing Fan, Richard Samworth and Yichao Wu Ultrahigh
  dimensional feature selection: beyond the linear model, 2009,
  Journal of Machine Learning Research 2013-2038.
  
  [6] Marius Hofert, Martin Machler, Alexander J. McNeil: Estimators
  for Archimedean copulas in high dimensions 2012-11-05,
  arXiv:1207.1708v2 [stat.CO] 2 Nov 2012
  
  [7] R.B. Nelsen. An Introduction to Copulas. Springer Science+
  Business Media, Inc., 2006.
  
  [8] Pekka Paalanen , Joni-Kristian Kamarainen, Jarmo Ilonen , Heikki
  Kelvininen, Feature representation and discrimination based on
  Gaussian mixture model probability densities practices and
  algorithms,Pattern Recognition, Volume 39, Issue 7, July 2006, Pages
  1346-1358
  
  [9] M Sklar. Fonctions de r'epartition `a n dimensions et leurs
  marges. Universit'e Paris 8, 1959.
  
  [10]  H. White. Estimation, Inference and Specification Analysis.
  Cambridge University Press, 1994
  
  [11] CF Jeff Wu. On the convergence properties of the em algorithm.
  The Annals of statistics, pages 95-103, 1983.
  
  [12] Lei Xu and Michael I Jordan. On convergence properties of the
  em algorithm for Gaussian mixtures. Neural computation,
  8(1):129-151, 1996.
  
  [13] Ming-Hsuan Yang, Narendra Ahuja: Gaussian mixture model for
  human skin color and its applications in image and video databases
  Proc. SPIE 3656, Storage and Retrieval for Image and Video Databases
  VII, 458 (December 17, 1998); doi:10.1117/12.333865
    
[14] Ke Wan and Alain Kornhauser: Turn-by-turn routing decision based on copula travel time estimation  with observable floating car data, TRB 2010

[15] Ke Wan: Estimation of travel time distribution and travel time derivatives, 2014

[16] Tewari, Ashutosh and Giering, Michael J and Raghunathan, Arvind:  Parametric characterization of multimodal distributions with non-gaussian modes, 2011 IEEE 11th international conference on data mining workshops, 2011

[17] Rajan, Vaibhav and Bhattacharya, Sakyajit: Dependency Clustering of Mixed Data with Gaussian Mixture Copulas, IJCAI, 2016

[18] Bilgrau, Anders Ellern and Eriksen, Poul Svante and Rasmussen, Jakob Gulddahl and Johnsen, Hans Erik and Dybk{\ae}r, Karen and B{\o}gsted, Martin: GMCM: Unsupervised clustering and meta-analysis using gaussian mixture copula models,  Journal of Statistical Software, 2016

[19] Kasa, Siva Rajesh and Bhattacharya, Sakyajit and Rajan, Vaibhav: Gaussian mixture copulas for high-dimensional clustering and dependency-based subtyping, Bioinformatics, 2020

[20] Sheikholeslami, Razi and Gharari, Shervan and Papalexiou, Simon Michael and Clark, Martyn P: VISCOUS: A Variance-Based Sensitivity Analysis Using Copulas for Efficient Identification of Dominant Hydrological Processes, Water Resources Research, 2021

[21] Feldman, Joseph and Kowal, Daniel R: Nonparametric Copula Models for Mixed Data with Informative Missingness, arXiv preprint arXiv:2210.14988, 2022

[22] Zou, Mingzhe and Holjevac, Ninoslav and {\DJ}akovi{\'c}, Josip and Kuzle, Igor and Langella, Roberto and Di Giorgio, Vincenzo and Djokic, Sasa Z: Bayesian CNN-BiLSTM and Vine-GMCM Based Probabilistic Forecasting of Hour-Ahead Wind Farm Power Outputs, IEEE Transactions on Sustainable Energy, 2022
  }
  
\end{document}